\documentclass{article}

\usepackage{arxiv}

\usepackage{amssymb}
\usepackage{amsmath}
\usepackage[utf8]{inputenc} 
\usepackage[T1]{fontenc}    
\usepackage{hyperref}       
\usepackage{url}            
\usepackage{booktabs}       
\usepackage{amsfonts}       
\usepackage{nicefrac}       
\usepackage{microtype}      
\usepackage{xcolor}         
\usepackage{algorithm}
\usepackage{algpseudocode}
\usepackage{graphicx}
\usepackage{multirow}
\usepackage[flushleft]{threeparttable}
\usepackage{float}
\usepackage{caption}
\usepackage{natbib}

\title{Less is More: Selective Reflection for Compatible and Efficient Knowledge Distillation in Large Language Models}

\author{
 Lingyuan Liu \\
  City University of Hong Kong \\
  \texttt{ly.liu@my.cityu.edu.hk} \\
   \And
 Mengxiang Zhang\footnote{$*$} \\
  The University of Hong Kong \\
  \texttt{mxzhang6@connect.hku.hk} \\
}
\begin{document}
\maketitle
\begin{abstract}
Knowledge Distillation (KD) is a fundamental technique for compressing large language models (LLMs) into compact, efficient student models. 
However, existing white-box KD methods mainly focus on balancing ground truth and student-generated responses while overlooking two critical factors: training data quality and student-model compatibility. 
To address these limitations, we propose Selective Reflection Distillation (SRD), a novel data curation framework that leverages reflections from student models to systematically refine training data.
SRD dynamically evaluates and selects prompt-response pairs by comparing ground truth data with student model outputs, selectively curating high-quality, student-compatible training instances through automated ranking based on difficulty.
Furthermore, after selecting the training data, a curriculum scheduling strategy is employed to incrementally introduce these curated subsets into the distillation process at fixed intervals.
As a plug-and-play enhancement, SRD consistently improves distillation outcomes across diverse white-box KD approaches and model architectures, as well as decreases computational cost significantly during KD training.
Experiments on a range of language model benchmarks demonstrate SRD’s consistent improvements in distilled model performance, as well as a reduction in training runtime by up to 39\%, under diverse KD methods and model families.
Notably, SRD operates as a plug-and-play module, enhancing sample efficiency without modifying underlying KD algorithms. 
Our findings highlight that data quality and compatibility are pivotal to effective and efficient distillation of LLMs, and SRD provides a principled framework to achieve both. This work advances the understanding of data-centric factors in KD and offers practical insights for enhancing the capability and efficiency of compressed LLMs.
\footnotetext{Corresponding author.}
\footnote{The code for our method is publicly available at \href{https://github.com/liuliuyuan6/SRD}{https://github.com/liuliuyuan6/SRD}.}
\end{abstract}

\section{Introduction}\label{sec:01}
\begin{figure}[t]
\centering
\includegraphics[width=0.91\textwidth]{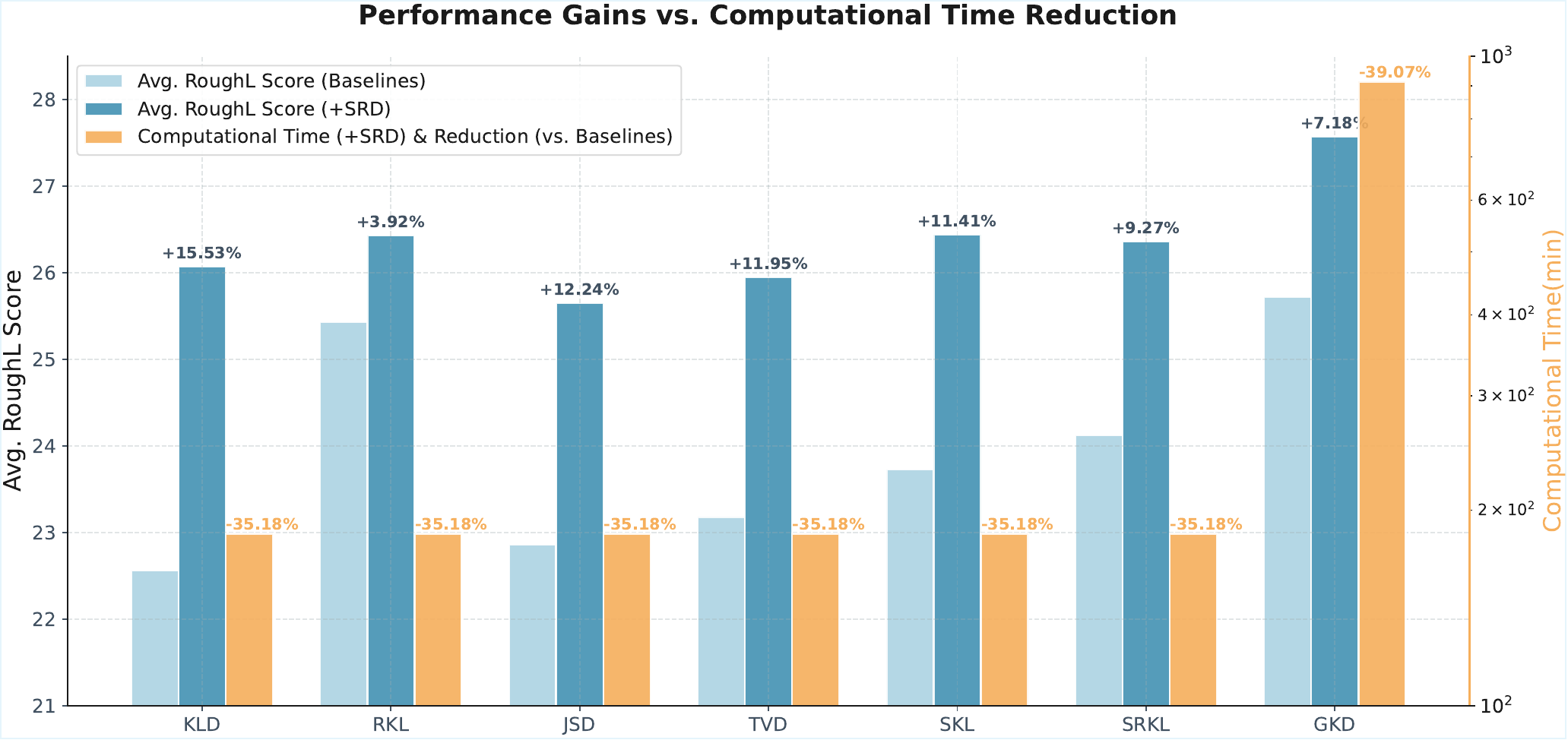} 
\caption{\footnotesize Effectiveness and Efficiency of SRD. Comparison of off-policy (KLD, RKL, JSD, TVD, SKL, SRKL) and on-policy (GKD) KD methods, with and without SRD, on five instruction-following benchmarks. \textbf{SRD consistently improves all baseline white-box KD methods substantially} in ROUGE-L scores while using 25\% less training data and \textbf{reducing runtime by 35\%} (OpenLLaMA2-7B $\rightarrow$ OpenLLaMA2-3B). This demonstrates SRD's dual advantage: enhanced performance through curated data selection and efficiency via reduced computational overhead.  }
\label{fig0}
\end{figure}

Large language models (LLMs) have achieved remarkable performance in text generation, language understanding, and reasoning, driven by scale and high-quality training data \citep{ouyang2022training}. However, their substantial computational and memory demands hinder deployment in resource-constrained environments such as edge devices \citep{aryan2023costly}. This has motivated the development of compact, efficient models that preserve strong performance across tasks like text generation \citep{li2024pre} and tool use \citep{gao2024confucius}. Knowledge distillation (KD) \citep{hinton2015distilling}, which transfers knowledge from a powerful teacher to a smaller student, has become a key technique for LLM compression.

KD methods for LLMs are broadly categorized as black-box or white-box \citep{yang2024survey}. Black-box KD uses only teacher outputs \citep{kim2016sequence}, making it suitable for proprietary models like GPT-4o \citep{hurst2024gpt} and Claude 3.5 \citep{claude35_sonnet_news}. It resembles data augmentation and curation \citep{xu2024survey}, where the teacher enhances training data quality. However, the rise of open-source LLMs—such as DeepSeek-v3 \citep{liu2024deepseek} and Qwen 2.5 \citep{yang2024qwen2}—with performance rivaling closed models \citep{AIIndex2025} has renewed interest in white-box KD. These methods leverage soft labels from the teacher’s probability distribution, offering richer supervision than one-hot labels.

In white-box KD, much effort has focused on designing divergence functions for the distillation loss. While Kullback-Leibler divergence (KLD) is widely used \citep{hinton2015distilling}, its asymmetry can lead to mode-averaging issues \citep{gu2023minillm}. Subsequent work has proposed alternatives such as reverse KLD (RKL) \citep{gu2023minillm}, Jensen-Shannon divergence (JSD) \citep{agarwal2024policy}, forward and reverse skew KLD (SKL and SRKL) \citep{ko2024distillm}, and $\alpha$-$\beta$-Divergence \citep{wang2025abkd}. Although effective in specific settings, these variants lack consistent generalization across tasks and datasets \citep{agarwal2024policy, ko2024distillm}, highlighting the need for complementary data strategies beyond loss design.

Another critical challenge in KD for LLMs is the training-inference mismatch arising from capacity gaps between teacher and student. To mitigate this, recent methods explore data curation strategies that combine ground truth outputs \citep{hinton2015distilling}, teacher generated outputs \citep{kim2016sequence}, and student generated outputs \citep{lin2020autoregressive, agarwal2024policy}. Student generated outputs-based approaches \citep{gu2023minillm, agarwal2024policy} improve compatibility and reduce mismatch but incur high computational costs and risk misguidance when low-quality student generated outputs dominate early training. Their success underscores a fundamental principle: data quality and student-model compatibility are central to effective distillation—a principle well-established in black-box KD \citep{ding2023enhancing} and supervised fine-tuning (SFT) \citep{li2023quantity, li2024selective}, yet often overlooked in white-box KD.

These observations reveal a core challenge in white-box KD: selecting what training data to use and when to introduce it during training to maximize student performance and reduce computational overhead. We identify two key questions: (1) What constitutes suitable training data? (2) What is the optimal timing for introducing such data? For the first, high-quality, student-compatible instances promote effective learning and reduce misguidance. For the second, the difficulty of training samples should align with the student’s evolving capacity.

\begin{figure}[t]
\centering
\includegraphics[width=0.7\textwidth]{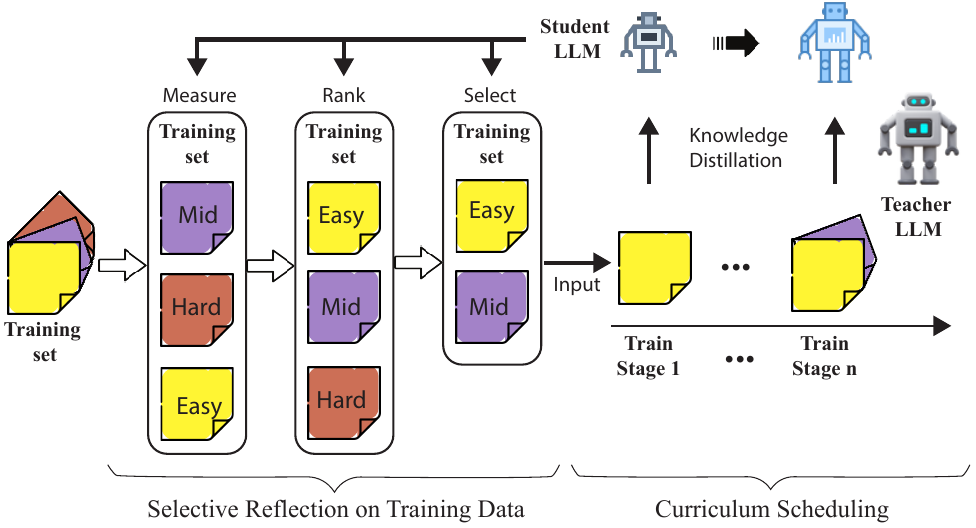} 
\caption{\footnotesize Overview of Selective Reflection Distillation (SRD). SRD is a two-stage framework for enhancing white-box KD. 1) \textbf{Selective Reflection on Training Data} measures sample difficulty using student generated outputs and ranks them via reciprocal rank fusion. The most difficult instances are eliminated to retain high-quality, student-compatible data. 2) \textbf{Curriculum Scheduling} partitions the curated data from easy to hard and incrementally introduces them during training, aligning the learning process with the student's evolving capacity.}
\label{fig1}
\end{figure}

To address these challenges, we propose \textbf{Selective Reflection Distillation (SRD)}, a novel data curation framework that leverages student model reflections to refine training data. To answer the first question, SRD dynamically evaluates prompt-response pairs using student confidence signals—ROUGE-L scores \citep{lin2004rouge} and cross-entropy loss—and ranks them via reciprocal rank fusion \citep{cormack2009reciprocal}. The most difficult instances are then eliminated based on a ratio, ensuring only high-quality, student-compatible data are retained. To address the second question, SRD partitions the curated data into difficulty-based subsets and employs a curriculum scheduling strategy to incrementally introduce them from easy to hard at fixed intervals (see Figure~\ref{fig1}). As a plug-and-play module, SRD reduces training data volume, cutting KD runtime by up to 39\%, while substantially improving distilled model performance. Empirical evaluations across instruction-following, mathematical reasoning, code generation, text summarization, and machine translation—using diverse model families including GPT-2 \citep{radford2019language}, OpenLLaMA2 \citep{touvron2023llama}, Qwen2.5 \citep{yang2024qwen2}, and T5 \citep{raffel2020exploring}—demonstrate SRD's generality, scalability, efficiency and effectiveness (See Fig. \ref{fig0}).

In summary, our contributions are two-fold:

\textbf{Methodologically}: We propose SRD, a plug-and-play data curation framework that selects high-quality, student-compatible training instances and introduces them progressively via curriculum scheduling, enhancing effectiveness, efficiency, and flexibility in white-box KD.

\textbf{Empirically}: Extensive experiments across multiple tasks, model architectures, and KD methods demonstrate that SRD consistently improves student performance while reducing computational cost, validating its generality and practical value.

\section{Background and Preliminary Study}\label{sec:02}

\subsection{Current Framework for White-Box KD}\label{subsec:0201}  
In white-box KD, a student language model (LM) learns from a more powerful teacher LM using a dataset of prompt-response pairs $(x, y)$. The student is trained to minimize two objectives: (1) the cross-entropy loss between the ground truth $y$ and the student's predicted distribution $q_{\theta}(y|x)$, and (2) the KD loss, which measures the divergence between the teacher’s soft-label distribution $p(y|x)$ and the student’s distribution $q_{\theta}(y|x)$. The cross-entropy loss is defined as:  
\begin{equation}\label{eq:ce}  
L_{ce} = -\sum_{i=1}^{|y|} \log q_{\theta}(y_i | x, y_{<i}),  
\end{equation}  
where $q_{\theta}(y_i | x, y_{<i})$ denotes the student's probability of generating token $y_i$ given input $x$ and previous tokens $y_{<i}$.

The KD loss is given by:  
\begin{equation}\label{eq:kd}  
L_{kd} = \sum_{i=1}^{|y|} D\left(p(y_i | x, y_{<i}; \tau) \parallel q_{\theta}(y_i | x, y_{<i}; \tau)\right),  
\end{equation}  
where $D(\cdot \parallel \cdot)$ is a divergence measure, and $\tau$ is the temperature parameter that softens the output distributions for smoother supervision.

The overall training objective combines both components:  
\begin{equation}\label{eq:total_loss}  
L_{s} = \alpha \cdot L_{ce} + (1 - \alpha) \cdot L_{kd},  
\end{equation}  
where $\alpha \in [0, 1]$ balances the contributions of ground truth fidelity and knowledge transfer.

\subsection{Challenges in Training Data Usage for White-Box KD}\label{subsec:0202}  
A persistent issue in KD for LLMs is the training-inference mismatch, caused by student-teacher capacity gaps and the static nature of training data \citep{ko2024distillm}. Reliance on ground truth outputs \citep{hinton2015distilling} often leads to ineffective transfer, as smaller students struggle to match the teacher's complex distributions. To address this, data curation strategies incorporate teacher generated outputs \citep{kim2016sequence} or, more effectively, using student generated outputs \citep{lin2020autoregressive, agarwal2024policy} to align training with inference dynamics, thereby improving compatibility and generalization.

However, student-generated output-based approaches face critical trade-offs. Early in training, low-quality student generated outputs from an underdeveloped student can introduce noise and misguide learning—a phenomenon known as teacher misguidance \citep{gu2023minillm}. Conversely, excessive student-generated output generation in later stages incurs high computational costs, often consuming up to 80\% of total training time \citep{ko2024distillm}. 

\subsection{Toward Principled Data Utilization in White-Box KD}\label{subsec:0203}  
The challenges and partial successes of student-generated output-based methods reveal a deeper principle: the effectiveness of KD hinges not only on \textit{what} data is used but also on \textit{when} it is introduced. This insight aligns with theoretical analyses suggesting that distillation dynamics are governed by the alignment of teacher and student output distributions \citep{wu2024rethinking, ko2025distillm}. For “easy” instances—where the student’s distribution $q_\theta(y|x)$ is initially close to the teacher’s $p(y|x)$—gradient updates are stable and efficient. In contrast, “hard” instances induce large distributional divergences, leading to high-variance gradients that can destabilize training, particularly for capacity-limited students.

From a data quality perspective, easy instances offer higher compatibility and lower noise, promoting smoother convergence. However, exclusive focus on such samples limits representational learning and underutilizes the teacher’s knowledge. Crucially, as the student improves during training, previously “hard” samples become more accessible—mirroring the progressive refinement observed in on-policy KD, where student generated outputs grow in quality over time.

This evolution suggests a curriculum-based approach: aligning the difficulty of training samples with the student’s evolving competence. By gradually introducing harder samples, the student can assimilate complex patterns without abrupt distributional shifts \citep{wang2021survey}. Such scheduling not only mimics the implicit curriculum of on-policy methods but also decouples data quality control from expensive student-generated output generation.

Therefore, we identify two foundational principles for effective data utilization in white-box KD:  
\begin{itemize}  
    \item \textbf{Difficulty-aware selection}: \textit{Prioritizing training instances where the student exhibits higher confidence—indicative of distributional alignment—leads to more stable and effective distillation. } 
    \item \textbf{Progressive learning}: \textit{Introducing training data in increasing order of difficulty enables controlled exposure to complex patterns, facilitating smoother optimization and better generalization, akin to human cognitive development. } 
\end{itemize}  
These principles form the theoretical foundation of our proposed SRD framework, which operationalizes them through student-informed data curation and structured curriculum design.

\section{Methodology}\label{sec:03}

\subsection{Motivation}\label{subsec:0301}  
Building on the insights from above Section, we propose Selective Reflection Distillation, a plug-and-play data curation framework that leverages student model reflections to improve the quality and training dynamics of white-box KD. SRD is designed for conditional text generation tasks, where a student language model learns from a fixed teacher model using a dataset of prompt-response pairs $(x, y)$. The framework enhances distillation by addressing two key questions: (1) \textit{What data should be used?} and (2) \textit{When should it be introduced?}

SRD consists of two core components:  

1. \textbf{Selective Reflection on Training Data}: Evaluates and filters training instances based on student model confidence, retaining only high-quality, student-compatible samples. 

2. \textbf{Curriculum Scheduling}: Progressively introduces curated data from easy to hard during training, aligning sample difficulty with the student’s evolving capacity.

SRD can integrate seamlessly into existing white-box KD pipelines and supports diverse divergence functions (e.g., KLD, JSD, SKL) and data types (e.g., ground truth outputs, student generated outputs). Algorithm~\ref{alg:cap} outlines the full procedure, with detailed descriptions provided in the following subsections.

\begin{algorithm}[tb]\small
\caption{SRD: Selective Reflection Distillation for LLM}\label{alg:cap}
\textbf{Input:} $D$: training dataset, $M_{s}$: student LM, $M_{t}$: teacher LM; 
\textbf{Output:} $M_{s}^{*}:$ the distilled student LM.
\begin{algorithmic}
\State $D'$ = sort($D$,$y_s$);
\State $\{\Delta_{1}, \Delta_{2}, ..., \Delta_{n} \}$ = $D''$ = filter($D'$,$\lambda$);
\State $D_{train} = \emptyset$, $\tau_0 = 1$, $\alpha_0 = 1$;
\State $M_{s}^{*}$ = $M_{s}$;
\For{$i = 1, ..., n $}
\State $D_{train} = D_{train} \bigcup \Delta_{i}$;
\State $\tau = \tau_i$, $\alpha = \alpha_i$;
    \While{}{not converged for $p$ epochs}
    \State $M_{s}^{*}$ = KD train($M_{s}^{*}$, $M_{t}$, $D_{train}$, $\tau_n$, $\alpha_n$);
    \EndWhile
\EndFor
\end{algorithmic}
\end{algorithm}

\subsection{Selective Reflection on Training Data}\label{subsec:0302}  
\textbf{Reflection on training data:}  
Given a prompt-response pair $(x, y)$ from the original dataset $D$ with $N$ samples, the student model generates a response $y_s$, forming a student-generated output. By comparing $y_s$ with the ground truth output $y$, we assess the sample's difficulty from the student's perspective. We use two complementary metrics:  
(1) ROUGE-L between $y$ and $y_s$: higher scores indicate better alignment and lower difficulty.  
(2) Cross-entropy loss over $y_s$ (Eq.~\ref{eq:ce}) : lower values reflect higher confidence and ease.

Each metric produces a ranked list—$\phi_{r}$ by ROUGE-L and $\phi_{e}$ by cross-entropy. These are combined using reciprocal rank fusion \citep{cormack2009reciprocal}, a robust method from information retrieval, to yield a unified difficulty ranking $\phi$. The fused score for each sample is:  
$f = \sum_{i}^{n} \frac{1}{k + r_{i}}$,  
where $r_i$ is the sample’s rank in list $i$, and $k = 60$ stabilizes the fusion. A higher $f$ indicates an easier, more suitable sample.

\textbf{Selection on training data:}  
Based on $\phi$, we eliminate the most difficult instances to ensure training focuses on high-confidence, student-compatible data. 
We retain only the top $\lambda$ fraction of samples, where $\lambda \in [0,1]$ is a configurable threshold that controls data retention (i.e., the hardest $1-\lambda$ fraction of samples are removed).
The curated dataset $D''$ is defined as:  
$D'' = \left\{ s_i \,\middle|\, i  \leq\lambda N \right\}$,  
where $s_i$ denotes the $i$-th ranked sample. This selective curation enhances data quality while reducing training burden.

\subsection{Curriculum Scheduling}\label{subsec:0303}  
After curating $D''$, SRD partitions it into $n$ subsets $\{\Delta_1, \dots, \Delta_n\}$ of increasing difficulty, each containing approximately $\frac{\lambda N}{n}$ or $\frac{\lambda N}{n} + 1$ samples. We adopt a discrete Baby Step curriculum \citep{bengio2009curriculum}, where training proceeds in $n$ stages. At stage $i$, the student is trained on the cumulative set $\bigcup_{j=1}^{i} \Delta_j$, starting with the easiest subset and incrementally adding harder ones until the full curated dataset is used. 

To further stabilize learning, SRD employs adaptive scheduling of key training hyperparameters. The distillation temperature $\tau$ (Eq.~\ref{eq:kd}) increases linearly from $\tau_0$ to $\tau_n$:  
$\tau_i = \tau_{0} + (\tau_{n} - \tau_{0}) \cdot \frac{i-1}{n-1}$,  
encouraging the student to first learn sharp, confident predictions before absorbing softer, more global teacher signals. The SFT ratio $\alpha$ in the total loss (Eq.~\ref{eq:total_loss}) decreases linearly from $\alpha_0$ to $\alpha_n$:  
$\alpha_i = \alpha_{0} - (\alpha_{0} - \alpha_{n}) \cdot \frac{i-1}{n-1}$,  
prioritizing ground truth fidelity early and knowledge transfer later—inspired by cognitive learning strategies \citep{libby2008comparison}. 

These dynamic schedules enhance training stability and performance while reducing reliance on computationally expensive student-generated output generation. Full details are provided in Algorithm~\ref{alg:cap}.

\section{Experiments}\label{sec:04}
We evaluate SRD across instruction-following, text summarization, machine translation, mathematical reasoning, and code generation tasks. SRD measures, ranks, and selects training instances based on student model reflections, retaining the top $\lambda = 75\%$ of samples by difficulty. The curated data are partitioned into $n = 3$ subsets, introduced incrementally under a Baby Step curriculum. Due to reduced data volume (training on 25\%, 50\%, and 75\% of the full dataset in successive stages), SRD reduces total training epochs to 60\% of the baseline. The distillation temperature $\tau$ increases linearly from $\tau_0 = 1$ to $\tau_n = 2$, while the SFT weight $\alpha$ decreases from $\alpha_0 = 0.3$ to $\alpha_n = 0.1$. We evaluate SRD across seven white-box KD baselines: GKD (on-policy) \citep{agarwal2024policy}, KLD \citep{hinton2015distilling}, RKL \citep{gu2023minillm}, JSD, TVD \citep{wen2023f}, SKL, and SRKL \citep{ko2024distillm}.

\begin{table*}[h]\tiny
    \centering
\begin{tabular}{ccl|cccccccc} \hline
$M_{t}$ & $M_{s}$ & KD Methods & \textbf{DollyEval} & \textbf{SelfInst} & \textbf{VicunaEval} & \textbf{S-NI} & \textbf{UnNI} & \textbf{Avg.} & \textbf{Performance Gains (\%)} & \textbf{Time Reduction (\%)} \\ \hline
        &            & $M_{t}$            & 26.26&19.15&17.31&31.20&31.84&25.15&-	   &-       \\\cline{3-11}     
        &            & SFT                & 24.87&18.78&16.50&28.65&28.73&23.51&-	   &-       \\       
        &            & GKD (on-policy)    & ~~26.30$^*$&18.91&~~17.53$^*$&~~34.21$^*$&31.67&~~25.72$^*$&-	   &  - \\       
        &            & ~~~~~~\tiny{+SRD}  & ~~\textbf{27.87}$^*$& ~~\textbf{19.74}$^*$&~~18.01$^*$&~~36.71$^*$&~~35.52$^*$&~~27.57$^*$&+7.18\% &-39.07\% \\       
        &            & KLD                & 23.76&17.48&15.68&28.42&27.48&22.56&-	   &  - \\       
        &            & ~~~~~~\tiny{+SRD}  & 25.81&18.44&~~17.56$^*$&~~34.44$^*$&~~34.09$^*$&~~26.07$^*$&+15.53\%&-35.18\% \\       
        &            & RKL                & ~~26.67$^*$&18.78&~~18.43$^*$&~~31.27$^*$&~~32.01$^*$&~~25.43$^*$&-	   &  - \\       
OpenLLaMA2&OpenLLaMA2& ~~~~~~\tiny{+SRD}  & 26.12&17.80&16.97&~~\textbf{35.65}$^*$&~~35.62$^*$&~~26.43$^*$&+3.92\% &-35.18\% \\       
(7B)    &(3B)        & JSD                & 25.64&17.78&15.56&27.93&27.39&22.86&-	   &  - \\       
        &            & ~~~~~~\tiny{+SRD}  & ~~26.79$^*$&17.99&17.01&~~33.27$^*$&~~33.20$^*$&~~25.65$^*$&+12.24\%&-35.18\% \\       
        &            & TVD                & 24.62&18.32&15.85&29.01&28.08&23.18&-	   &  - \\       
        &            & ~~~~~~\tiny{+SRD}  & ~~26.79$^*$&~~17.69$^*$&17.69&~~34.37$^*$&~~33.19$^*$&~~25.95$^*$&+11.95\%&-35.18\% \\       
        &            & SKL                & 24.99&18.77&16.69&29.71&28.49&23.73&-	   &  - \\       
        &            & ~~~~~~\tiny{+SRD}  & ~~26.57$^*$&17.64&~~18.11$^*$&34.60&~~\textbf{35.27}$^*$&~~26.44$^*$&+11.41\%&-35.18\% \\       
        &            & SRKL               & 25.67&19.11&16.93&30.07&28.81&24.12&-	   &  - \\       
        &            & ~~~~~~\tiny{+SRD}  & ~~26.99$^*$&18.37&~~\textbf{18.55}$^*$&~~35.30$^*$&~~32.57$^*$&~~26.36$^*$&+9.27\% &-35.18\% \\\hline    
        &       & $M_{t}$            &27.19&14.04&16.47&27.66&31.86&23.44&-	    &-      \\\cline{3-11}     
        &       & SFT                &23.33&10.01&14.72&16.38&19.57&16.80&-	    &-      \\       
        &       & GKD (on-policy)    &24.67&11.48&15.66&23.81&25.26&20.18&-     &  -\\       
        &       & ~~~~~~\tiny{+SRD}  &24.55&11.76&15.77&26.54&27.53&21.19&+5.23\%&-37.92\%\\       
        &       & KLD                &23.49&10.33&14.96&19.71&22.01&18.10&-	    &  -\\       
        &       & ~~~~~~\tiny{+SRD}  &24.59&11.07&15.15&20.75&23.25&18.96&+4.77\%&-33.53\%\\       
        &       & RKL                &23.79&12.13&14.94&23.81&22.52&19.44&-     &  -\\       
GPT-2   &GPT-2  & ~~~~~~\tiny{+SRD}  &24.50&12.06&15.03&24.13&24.03&19.95&+2.64\%&-33.53\%\\       
(1.5B)  &(0.1B) & JSD                &24.07&11.38&15.87&22.84&23.06&19.44&-	    &  -\\       
        &       & ~~~~~~\tiny{+SRD}  &24.85&11.00&15.93&25.99&24.69&20.45&+5.39\%&-33.53\%\\       
        &       & TVD                &24.32&11.09&15.51&25.93&26.55&20.68&-     &  -\\       
        &       & ~~~~~~\tiny{+SRD}  &24.55&12.39&15.80&~~\textbf{27.90}$^*$&28.88&21.80&+5.91\%&-33.53\%\\       
        &       & SKL                &24.24&12.27&15.71&23.33&24.02&19.91&-     &  -\\       
        &       & ~~~~~~\tiny{+SRD}  &25.26&12.72&15.74&26.20&27.79&21.48&+8.18\%&-33.53\%\\       
        &       & SRKL               &25.22&12.86&15.18&25.51&28.43&21.44&-     &  -\\       
        &       & ~~~~~~\tiny{+SRD}  &\textbf{25.41}&\textbf{13.17}&\textbf{15.92}&~~27.88$^*$&\textbf{29.19}&22.31&+4.07\%&-33.53\%\\\hline
\end{tabular}
\caption{\footnotesize Instruction-following evaluation on OpenLLaMA2 and GPT-2. $M_t$ and $M_s$ denote teacher and student models. Baselines include SFT and seven white-box KD methods: GKD (on-policy, using 50\% student generated outputs), and off-policy variants (KLD, RKL, JSD, TVD, SKL, SRKL). \textbf{DollyEval}, \textbf{SelfInst}, \textbf{VicunaEval}, \textbf{S-NI}, and \textbf{UnNI} report Rouge-L scores; \textbf{Avg.} is the mean across datasets. Best student results are \textbf{bold}; $^*$ indicates student outperforms teacher. Results are averaged over five random seeds.}
\label{tab1}
\end{table*}

\subsection{General Instruction-Following}\label{sec:0401}  
\subsubsection{Setup.}\label{sec:040101}  
We evaluate SRD using two model families: GPT-2 (0.1B student, 1.5B teacher) and OpenLLaMA2 (3B student, 7B teacher). Training uses the Databricks-dolly-15k dataset \citep{DatabricksBlog2023DollyV2}, filtered and split into 11.5K train, 1K validation, and 0.5K test samples. Hyperparameters are tuned on the validation set using Rouge-L \citep{lin2004rouge}. Evaluation is conducted on five instruction-following benchmarks: DollyEval, SelfInst, VicunaEval, S-NI, and UnNI. We report average Rouge-L scores over five random seeds, with generation temperature set to 1.

\subsubsection{Results.}\label{sec:040102} 
As shown in Table~\ref{tab1}, SRD consistently improves student performance across all KD methods and model scales. SRD achieves significant gains over larger language models, with the distilled student models not only surpassing their respective baseline performances but also exceeding the teacher model on multiple evaluation datasets. The performance improvements are pronounced under various divergence-based distillation objectives, demonstrating the compatibility and effectiveness of SRD across different loss formulations. For both model families, SRD enables the student to achieve the highest performance when combined with advanced KD losses, highlighting its synergistic interaction with modern distillation techniques. All improvements are statistically significant under the Wilcoxon signed-rank test ($p < 0.05$). Despite using only 75\% of the training data and 60–67\% of baseline training time, SRD delivers superior performance, demonstrating its effectiveness, efficiency, and generalization across model architectures and distillation strategies.

\subsection{Text Summarization and Machine Translation}\label{sec:0402}  
\begin{table}[h]\scriptsize
    \centering
\begin{tabular}{l|cccc} \hline
            &  XSum & IWSLT & GSM8K & MBPP \\
KD Methods & ROUGE-L & BLEU & \textit{Pass@1} & \textit{Pass@1} \\ \hline
 $M_{t}$            &30.86&34.56&89.42&74.63\\ \hline
 GKD (on-policy)    &27.99&30.24&80.21&61.58\\
 ~~~~~~\tiny{+SRD}  &\textbf{29.24}&\textbf{31.18}&80.58&61.93\\
 KLD                &27.15&29.36&77.92&61.06\\
 ~~~~~~\tiny{+SRD}  &26.68&30.29&78.21&61.43\\
 RKL                &28.06&29.54&78.39&61.37\\
 ~~~~~~\tiny{+SRD}  &28.88&30.26&79.67&61.54\\
 JSD                &27.24&29.73&78.92&61.00\\
 ~~~~~~\tiny{+SRD}  &27.97&30.77&79.17&61.91\\
 TVD                &27.34&28.95&78.19&61.65\\
 ~~~~~~\tiny{+SRD}  &27.49&29.36&79.10&\textbf{62.00}\\
 SKL                &27.52&29.84&79.43&61.57\\
 ~~~~~~\tiny{+SRD}  &28.50&30.83&\textbf{80.77}&61.76\\
 SRKL               &27.64&30.01&79.49&61.87\\
 ~~~~~~\tiny{+SRD}  &28.74&31.09&82.33&62.75\\ \hline  
\end{tabular}
\caption{\footnotesize Performance comparison on XSum (summarization), IWSLT (machine translation), GSM8K (mathematical reasoning), and MBPP (code generation). Metrics: ROUGE-L (XSum), BLEU (IWSLT), and \textit{Pass@1} (GSM8K, MBPP). Teacher-student pairs: T5-XL (3B) $\rightarrow$ T5-Base (0.2B) for XSum; mT5-XL (3B) $\rightarrow$ mT5-Base (0.2B) for IWSLT; Qwen2.5-Math-7B $\rightarrow$ Qwen2.5-Math-1.5B for GSM8K; Qwen2.5-Coder-7B $\rightarrow$ Qwen2.5-Coder-1.5B for MBPP. Best student results are \textbf{bold}.}
\label{tab2}
\end{table}

\subsubsection{Setup.}\label{sec:040201}  
We evaluate SRD on text summarization using the XSum dataset \citep{narayan2018don}, which contains single-document, abstractive summaries. The student is T5-Base (0.2B) fine-tuned on XSum, with T5-XL (3B) as the teacher. For machine translation, we use the IWSLT 2017 en-de dataset \citep{cettolo2017overview}, evaluating translation quality via BLEU \citep{papineni2002bleu}. The student is mT5-Base v1.1 (0.2B), and the teacher is mT5-XL (3B). 

\subsubsection{Results.}\label{sec:040202}  
As shown in Table~\ref{tab2}, SRD consistently enhances student performance across both summarization and machine translation tasks under all KD methods. The improvements are statistically significant and align with gains observed in instruction-following, confirming that SRD’s effectiveness generalizes to tasks where token-level similarity (ROUGE-L, BLEU) correlates well with output quality. 

\subsection{Mathematical Reasoning and Code Generation}\label{sec:0403}  
\subsubsection{Setup.}\label{sec:040301}  
For mathematical reasoning, we use GSM8K \citep{cobbe2021training}, a benchmark of multi-step grade school math problems. The teacher is Qwen2.5-Math-7B-Inst and the student is Qwen2.5-Math-1.5B, both fine-tuned on MetaMathQA \citep{yu2023metamath}. We employ chain-of-thought prompting \citep{wei2022chain} to elicit step-by-step reasoning. For code generation, we use prompts from the WizardCoder dataset \citep{luo2023wizardcoder}, derived via EvolInstruct \citep{xu2024wizardlm}. The teacher is Qwen2.5-Coder-7B-Inst and the student is Qwen2.5-Coder-1.5B. Evaluation is performed on MBPP \citep{austin2021program} using \textit{Pass@1}.

\subsubsection{Results.}\label{sec:040302}  
SRD yields positive but more moderate improvements on mathematical reasoning and code generation, as shown in Table~\ref{tab2}. This suggests that while the framework is broadly applicable, the effectiveness decreases in domains where correctness is highly sensitive to minor errors—such as a single miscalculation or syntax mistake—that are not well captured by ROUGE-L or cross-entropy. These metrics may overestimate similarity despite functional failure. These findings suggest that its performance can be further enhanced by task-adaptive difficulty metrics. For example, future versions could incorporate execution-based feedback (e.g., code compilation or math verification) or semantic equivalence scoring for non-verbatim correctness.

\section{Analysis and Discussion}\label{sec:05}
To understand the effectiveness of SRD, we conduct ablation studies on the Dolly dataset using GPT-2 and OpenLLaMA2 model families. We analyze the contributions of SRD’s core components, evaluate the sensitivity of its hyperparameters, and compare its design choices against other curriculum-based methods. 

\subsection{Impact of Selective Reflection via Difficulty Ranking}\label{subsec:0501}  
\begin{table}[h]\scriptsize
    \centering
\begin{tabular}{l|ccccc} \hline
KD Methods & \textbf{DollyEval} & \textbf{SelfInst} & \textbf{VicunaEval} & \textbf{S-NI} & \textbf{UnNI} \\ \hline
 KLD                        & 23.49&10.33&14.96&19.71&22.01\\
 ~~~~~~\tiny{+SRD} (Fusion) & 24.59&11.07&15.15&20.75&23.25\\
 ~~~~~~\tiny{+SRD} (RoughL)    & 24.26&10.73&14.92&19.66&22.27\\
 ~~~~~~\tiny{+SRD} (Loss)   & 23.44&10.61&14.98&20.57&22.80\\ \hline
\end{tabular}
\caption{\footnotesize Ablation on ranking strategies within SRD. Performance of KLD with SRD using different difficulty estimation methods: fusion of ROUGE-L and cross-entropy (Fusion), ROUGE-L only (RoughL), and cross-entropy only (Loss). Teacher: GPT-2 (1.5B), Student: GPT-2 (0.1B). All scores are Rouge-L.}
\label{tab3}
\end{table}

The Selective Reflection component of SRD relies on accurate difficulty estimation to curate student-compatible training data. We compare three ranking strategies: using ROUGE-L scores only, cross-entropy loss only, and a fused ranking via reciprocal rank fusion. As shown in Table~\ref{tab3}, the fused ranking consistently yields the best performance across all evaluation datasets. While both single-metric variants improve over the baseline KLD, they underperform compared to the fusion approach, confirming that combining complementary signals enhances the reliability of difficulty assessment. This validates the design of SRD’s reflection mechanism and highlights the importance of multi-metric evaluation in data curation for KD.

\subsection{Role of Adaptive Parameter Scheduling in Curriculum Learning}\label{subsec:0502}  
\begin{table}[h]\scriptsize
    \centering
\begin{tabular}{l|ccccc} \hline
KD Methods & \textbf{DollyEval} & \textbf{SelfInst} & \textbf{VicunaEval} & \textbf{S-NI} & \textbf{UnNI} \\ \hline
 KLD                                       & 23.49&10.33&14.96&19.71&22.01\\
 ~\tiny{+SRD}                         & 24.59&11.07&15.15&20.75&23.25\\
 ~\tiny{+SRD (w/o T \& R)}  & 23.42&10.40&14.78&18.27&21.31\\
 ~\tiny{+SRD (w/o T) }            & 21.73&8.57 &14.58&18.95&21.61\\ 
 ~\tiny{+SRD (w/o R) }            & 23.85&10.97&15.04&20.28&23.06\\\hline
\end{tabular}
\caption{\footnotesize Ablation on adaptive parameter scheduling. “w/o T \& R”: no temperature and SFT ratio scheduling; “w/o T”: fixed temperature; “w/o R”: fixed SFT ratio. Teacher: GPT-2 (1.5B), Student: GPT-2 (0.1B).}
\label{tab4}
\end{table}

Curriculum scheduling in SRD includes adaptive control of the distillation temperature $\tau$ and the SFT ratio $\alpha$. We evaluate the impact of removing these dynamic schedules. As shown in Table~\ref{tab4}, disabling both schedules leads to marginal gains, while fixing only the temperature results in significant performance degradation. In contrast, fixing only the SFT ratio has a milder effect. This indicates that increasing the temperature over stages—starting with sharp, confident targets and progressing to softer teacher distributions—is a key driver of SRD’s effectiveness. 

\subsection{Influence of Training Order and Scheduling Direction}\label{subsec:0503}  
\begin{table}[h]\scriptsize
    \centering
\begin{tabular}{l|ccccc} \hline
KD Methods & \textbf{DollyEval} & \textbf{SelfInst} & \textbf{VicunaEval} & \textbf{S-NI} & \textbf{UnNI} \\ \hline
 KLD                         & 23.49&10.33&14.96&19.71&22.01\\
 ~\tiny{+SRD (Standard)}                & 24.59&11.07&15.15&20.75&23.25\\
 ~\tiny{+SRD (Reverse)}          & 23.42&10.40&14.78&18.27&21.31\\
 ~\tiny{+SRD (T:2-1) }       & 21.73&8.57 &14.58&18.95&21.61\\ 
 ~\tiny{+SRD (R:0.1-0.3) }     & 23.85&10.97&15.04&20.28&23.06\\\hline
\end{tabular}
\caption{\footnotesize Ablation on training order and scheduling direction. Standard: easy-to-hard sample order, $\tau$ from 1 to 2, $\alpha$ from 0.3 to 0.1. Reverse: hard-to-easy order. T:2-1: temperature from 2 to 1. R:0.1-0.3: SFT ratio from 0.1 to 0.3. Teacher: GPT-2 (1.5B), Student: GPT-2 (0.1B).}
\label{tab5}
\end{table}

\begin{figure*}[t]
\centering
\includegraphics[width=1\textwidth]{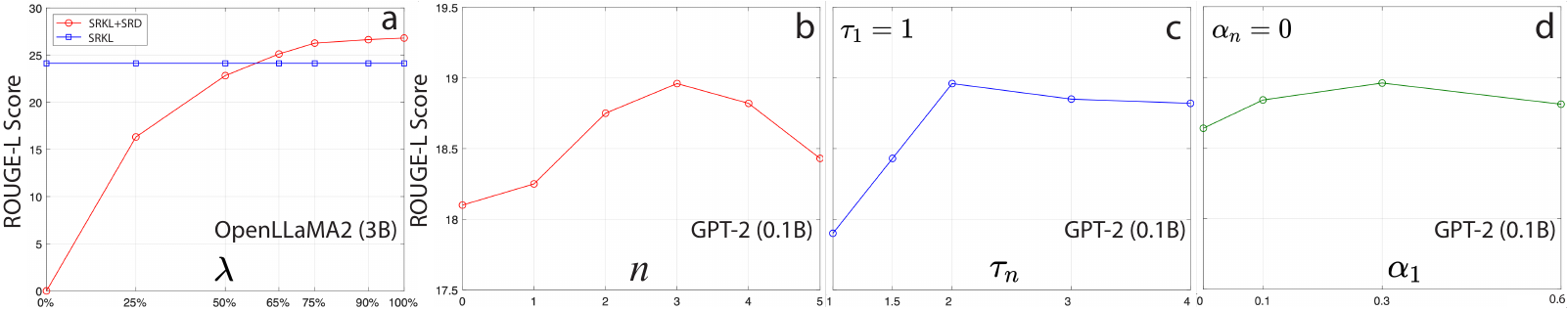}
\caption{\footnotesize Hyperparameter sensitivity analysis for SRD. (a) Performance vs. data retention ratio ($\lambda$); (b) Performance vs. number of curriculum stages ($n$); (c) Performance vs. final distillation temperature ($\tau_n$); (d) Performance vs. initial SFT ratio ($\alpha_0$). All results are averaged across five evaluation datasets using appropriate teacher-student model pairs and KD methods.}
\label{fig3}
\end{figure*}

We examine the importance of training order and scheduling direction. As shown in Table~\ref{tab5}, an easy-to-hard curriculum outperforms a hard-to-easy alternative, confirming that aligning sample difficulty with the student’s learning trajectory enhances knowledge transfer. Starting with complex samples hinders convergence and limits final performance. Similarly, reversing the temperature schedule (from high to low) severely degrades results, as early exposure to soft, uncertain teacher outputs destabilizes learning. In contrast, a decreasing SFT ratio (prioritizing ground truth early) proves beneficial, consistent with cognitive learning principles. These findings reinforce that SRD’s design—progressive difficulty, rising temperature, and fading supervision—is critical for optimal distillation dynamics.

\subsection{Hyperparameter Sensitivity Analysis}\label{subsec:0503}  
SRD introduces four key hyperparameters: the data retention ratio ($\lambda$), number of curriculum stages ($n$), distillation temperature range ($\tau_0$ to $\tau_n$), and SFT ratio range ($\alpha_0$ to $\alpha_n$). Figure~\ref{fig3} presents a comprehensive sensitivity analysis of these parameters across model families and KD methods.

\textbf{Data retention ratio} ($\lambda$): Performance increases with higher $\lambda$ values due to greater data availability, but exhibits diminishing returns beyond a threshold. This indicates that while a sufficient volume of training data is beneficial, the most difficult samples contribute little to effective knowledge transfer, supporting the rationale for selective removal of high-difficulty instances.

\textbf{Number of curriculum stages} ($n$): Performance is sensitive to the number of stages, with an optimal value that balances structured progression and training stability. Too few stages limit the curriculum effect, while too many disrupt learning continuity, suggesting that a moderate number of well-separated difficulty levels maximizes distillation efficacy.

\textbf{Final distillation temperature} ($\tau_n$): The final temperature in the increasing schedule has a non-linear impact on performance, with an optimal range that enables the student to first learn sharp, confident predictions before gradually absorbing softer, more informative teacher signals. This confirms the importance of progressive exposure to probabilistic guidance.

\textbf{Initial SFT ratio} ($\alpha_0$): An optimal initial value exists that prioritizes ground truth alignment in early stages without overly suppressing knowledge transfer later. Setting $\alpha_0$ too high prolongs reliance on supervision, while setting it too low undermines early stability, demonstrating the need for a balanced, decaying supervision schedule.

\subsection{Comparison with Traditional Curriculum Learning}\label{subsec:0504}  
\begin{table}[h]\scriptsize
    \centering
\begin{tabular}{l|cc|cc} \hline
& OpenLLaMA2 & (7B/3B) & GPT-2 & (1.5B/0.1B) \\
KD Methods & \textbf{Avg.} & \textbf{Gains (\%)} & \textbf{Avg.} & \textbf{PG (\%)} \\ \hline
 KLD                   & 22.56&-        & 18.10&-      \\
 ~\tiny{+SRD}          & 26.07&15.53\%  & 18.96&4.78\% \\
 ~\tiny{+TCL}          & 22.91&1.53\%   & 17.91&1.01\%\\ \hline
\end{tabular}
\caption{\footnotesize Comparison of SRD with TCL on instruction-following tasks. Models: OpenLLaMA2 (7B$\rightarrow$3B) and GPT-2 (1.5B$\rightarrow$0.1B), using KLD as the distillation loss. \textbf{Avg.} is the average ROUGE-L score across five datasets. \textbf{Gains} is the performance gains compares each method to its baseline. TCL uses word and sentence entropy- and length-based difficulty scoring without data curation.}
\label{tab6}
\end{table}

We compare SRD’s curriculum scheduling against a traditional curriculum learning (TCL) approach \citep{zhu2021combining}, which uses handcrafted features—such as token entropy and sequence length—to estimate sample difficulty and applies a fixed curriculum without data filtering. As shown in Table~\ref{tab6}, TCL yields marginal or even negative gains, while SRD achieves substantial improvements across both model families. This highlights the limitations of static, proxy-based difficulty metrics in modern KD pipelines. In contrast, SRD's student-informed reflection mechanism—combining dynamic difficulty assessment, selective data curation, and progressive learning—proves significantly more effective. These results demonstrate that SRD is not only compatible with white-box KD but also superior to conventional curriculum learning methods, emphasizing the importance of data quality, student-model compatibility, and adaptive training design in KD for LLMs.

\section{Related Work}\label{sec:06}
White-box KD for LLMs improves upon black-box approaches by leveraging soft labels from teacher models \citep{yang2024survey}, with extensive research devoted to designing divergence-based loss functions and incorporating teacher generated outputs or student generated outputs to enhance training dynamics \citep{kim2016sequence, agarwal2024policy}. However, these methods often overlook the critical importance of data quality and student-model compatibility—principles well recognized in black-box KD and SFT \citep{li2023quantity, li2024selective, ding2023enhancing}. While approaches like UltraChat \citep{ding2023enhancing} and phi \citep{gunasekar2023textbooks} emphasize high-quality data curation in black-box settings, and curriculum learning has been explored in vision and limited NLP tasks \citep{xiang2020learning, li2023curriculum, zhu2021combining}, extensions such as MPDistil \citep{sengupta2023good}, Confucius \citep{gao2024confucius}, and POCL \citep{liu2025being} still fail to integrate difficulty-aware data selection with student-centric curriculum design. This gap—between static data usage and dynamic learning needs—remains underaddressed in white-box KD, motivating our focus on principled, compatibility-driven data curation.

\section{Conclusion}\label{sec:07}
In this work, we present \textbf{Selective Reflection Distillation (SRD)}, a novel data curation framework for white-box KD in LLMs. Inspired by the established principle in SFT that data quality and student-model compatibility are critical to effective learning, SRD leverages student model reflections—via generated responses and confidence signals—to systematically identify and retain high-quality, student-compatible training instances. This addresses the fundamental question in KD: \textit{what data should be used?}
Then, SRD employs a curriculum scheduling strategy that incrementally introduces curated data from easy to hard at fixed intervals, aligning training progression with the student’s evolving capacity. This answers the second key question: \textit{when should data be introduced?} 
As \textbf{a plug-and-play enhancement}, SRD seamlessly integrates into existing white-box KD pipelines without modifying model architectures or loss functions. \textbf{It reduces knowledge distillation runtime by up to 39\%, while significantly enhancing distilled model performance} across diverse tasks, model families, and distillation methods. Our experiments demonstrate SRD’s generality, scalability, effectiveness, and efficiency, making it a practical solution for real-world LLM compression.
For future work, we plan to explore task-specific difficulty metrics and adaptive thresholds for sample elimination, further enhancing SRD’s flexibility and performance.

\appendix
\section{Preliminary Formulation of White-box KD}
We provide a formal overview of white-box KD for autoregressive language models, where a student model learns not only from ground-truth labels but also from the soft output distributions of a pre-trained teacher model. Consider a training set consisting of prompt-response pairs $(x, y)$, where $x$ denotes the input prompt and $y = (y_1, \dots, y_{|y|})$ the target response sequence. The student model, parameterized by $\theta$, is trained to minimize a composite objective that combines supervised SFT and KD.

The SFT component is defined via the standard cross-entropy loss:

\begin{equation}\label{eq:ce}
L_{ce} = -\sum_{i=1}^{|y|} \log q_{\theta}(y_i | x, y_{<i}),
\end{equation}

where $q_{\theta}(y_i | x, y_{<i})$ represents the conditional probability assigned by the student to token $y_i$, given the context $x$ and preceding tokens $y_{<i}$. This probability is computed using a softmax over the student's raw logits $z^s$:

\begin{equation}
q_{\theta}(y_i | x, y_{<i}) = \frac{\exp(z^s_{y_i})}{\sum_{j \in V} \exp(z^s_j)},
\end{equation}

with $z^s_{y_i}$ denoting the logit for token $y_i$ and $V$ the vocabulary set.

In parallel, the KD objective encourages the student to mimic the teacher model’s per-token output distribution. The distillation loss is expressed as:

\begin{equation}\label{eq:kd}
L_{kd} = -\tau^2 \cdot \sum_{i=1}^{|y|} D\left(p(y_i | x, y_{<i}; \tau) \parallel q_{\theta}(y_i | x, y_{<i}; \tau)\right),
\end{equation}

where $D(\cdot \parallel \cdot)$ denotes a statistical divergence—commonly Kullback–Leibler divergence (KLD) or Jensen–Shannon divergence (JSD)—between the temperature-scaled distributions of the teacher $p$ and student $q$. The temperature parameter $\tau > 0$ controls the smoothness of the output distribution, enabling softer probability assignments that convey richer information than one-hot labels.

The temperature-scaled probabilities are computed as:

\begin{equation}\label{eq:tem_q}
q_{\theta}(y_i | x, y_{<i}; \tau) = \frac{\exp(z^s_{y_i}/\tau)}{\sum_{j \in V} \exp(z^s_j/\tau)}
\end{equation}
and
\begin{equation}\label{eq:tem_p}
p(y_i | x, y_{<i}; \tau) = \frac{\exp(z^t_{y_i}/\tau)}{\sum_{j \in V} \exp(z^t_j/\tau)},
\end{equation}
where $z^t_j$ denotes the logits produced by the teacher model. The $\tau^2$ scaling factor in $L_{kd}$ ensures gradient magnitude consistency between $L_{ce}$ and $L_{kd}$ as $\tau$ varies, following the standard practice in distillation literature \citep{hinton2015distilling}.

The overall training objective combines both components:

\begin{equation}\label{eq:total_loss}
L_s = \alpha \cdot L_{ce} + (1 - \alpha) \cdot L_{kd},
\end{equation}
where $\alpha \in [0, 1]$ governs the trade-off between fidelity to ground-truth labels and alignment with the teacher’s softened outputs.

Existing research in white-box KD has explored various aspects of this framework, particularly the choice of divergence measure $D(\cdot \parallel \cdot)$, including vanilla KLD \citep{hinton2015distilling}, reverse KLD \citep{gu2023minillm}, JSD \citep{agarwal2024policy}, and skewed variants \citep{ko2024distillm}. Additionally, alternative data curation strategies have been investigated, such as training on human-annotated responses \citep{hinton2015distilling}, teacher-generated outputs (TGOs) \citep{kim2016sequence}, or student-generated outputs (SGOs) \citep{lin2020autoregressive, agarwal2024policy}, as well as hybrid approaches combining multiple sources \citep{gu2023minillm, ko2024distillm, ko2025distillm}. These efforts highlight the importance of data quality and source selection in effective knowledge transfer—motivating our proposed Selective Reflection Distillation framework.

\section{Detailed Experimental Setup}
We provide a comprehensive description of the experimental setup, including the base models, datasets, training procedures, evaluation protocols, and baseline methods used in our study.

\subsection{Base Models Description}
We conduct experiments across four prominent families of LLMs—GPT-2, OpenLLaMA2, Qwen2.5, and T5—spanning diverse architectures and scales to ensure broad applicability of our findings. These models were selected for their open availability, architectural representativeness, and widespread use in downstream evaluation benchmarks, facilitating reproducibility and comparative analysis.

\begin{itemize}
    \item \textbf{GPT-2} \citep{radford2019language}: A decoder-only transformer architecture introduced by OpenAI. In our knowledge distillation (KD) framework, we employ GPT-2 (1.5B) as the teacher and GPT-2 (0.1B) as the student model, focusing on instruction-following capabilities. This configuration enables investigation of performance transfer within the same architectural family under significant parameter disparity.

    \item \textbf{OpenLLaMA2} \citep{touvron2023llama}: An openly licensed reimplemention of Meta’s LLaMA architecture, trained on similar data and objectives. We use OpenLLaMA2 (7B) as the teacher and OpenLLaMA2 (3B) as the student for instruction-following tasks, allowing us to evaluate distillation efficacy in models with permissive licensing and community-driven development.

    \item \textbf{Qwen2.5} \citep{team2024qwen2}: A suite of specialized LLMs developed by the Qwen team, designed for domain-specific reasoning. For mathematical reasoning, we use Qwen2.5-Math-7B-Inst as the teacher and Qwen2.5-Math-1.5B as the student; for code generation, Qwen2.5-Coder-7B-Inst serves as the teacher and Qwen2.5-Coder-1.5B as the student. These pairings allow targeted evaluation of distillation in vertically specialized models.

    \item \textbf{T5} \citep{raffel2020exploring}: A text-to-text transformer model developed by Google, where all NLP tasks are framed as text generation. For text summarization, we use T5-Base (0.2B) as the student and T5-XL (3B) as the teacher. For multilingual machine translation, we adopt mT5-Base v1.1 (0.2B) and mT5-XL (3B) \citep{xue2020mt5}, which were pretrained on a multilingual corpus covering 101 languages derived from Common Crawl. This setup enables assessment of distillation in both monolingual and multilingual sequence-to-sequence scenarios.
\end{itemize}

All models are initialized with publicly available checkpoints, and fine-tuning is conducted under consistent optimization settings to ensure fair comparison.

\subsection{Dataset Description}
We evaluate our proposed Selective Reflection Distillation (SRD) framework across five core NLP tasks: instruction-following, text summarization, machine translation, mathematical reasoning, and code generation. Below, we describe the datasets used for training and evaluation in each domain.

\begin{itemize}
    \item \texttt{databricks-dolly-15K} \citep{DatabricksBlog2023DollyV2}: A diverse collection of 15,000 human-authored instruction-response pairs spanning seven categories, including brainstorming, closed QA, and summarization. This dataset provides high-quality, naturally occurring prompts for training and evaluating instruction-following behavior.

    \item \texttt{self-instruct-eval} \citep{wang2022self}: A synthetic data generation framework that produces instruction-following data via bootstrapping from model outputs. We use its training split (52,000 instructions, 82,000 input-output pairs) and evaluation set (252 expert-curated tasks and 50,000 public examples) to assess generalization to unseen task formulations.

    \item \texttt{vicuna-eval} \citep{chiang2023vicuna}: A challenging benchmark consisting of 80 complex, open-ended questions requiring multi-step reasoning and nuanced understanding. It is widely used to evaluate chat-based models and serves as a rigorous test for instruction adherence and reasoning depth.

    \item \texttt{supernatural-instructions} \citep{wang2022benchmarking}: A benchmark with 1,616 expert-written NLP tasks across 76 types. The test set includes approximately 9,000 samples from 119 tasks, enabling evaluation of zero-shot generalization across diverse task families.

    \item \texttt{unnatural-instructions-core} \citep{honovich2022unnatural}: A synthetically generated dataset containing 66,000 instruction-response pairs created through automated perturbation of natural instructions. The core subset is used to examine model performance when trained on large-scale machine-generated data.

    \item \texttt{XSum} \citep{gliwa2019samsum}: A dataset of over 200,000 BBC news articles, each paired with a single-sentence abstractive summary. It emphasizes extreme summarization, where the model must extract the central idea concisely, making it suitable for evaluating factual consistency and salience detection.

    \item \texttt{IWSLT 2017} \citep{cettolo2017overview}: A multilingual corpus for spoken language translation. We focus on the English-to-German (En-De) translation direction, using standard preprocessing and tokenization pipelines to align with common practice in machine translation benchmarks.

    \item \texttt{GSM8K} \citep{cobbe2021training}: A dataset of 8,500 linguistically varied grade school math word problems requiring multi-step arithmetic reasoning. Each problem demands coherent chain-of-thought decomposition, making it a standard benchmark for evaluating reasoning fidelity in distilled models.

    \item \texttt{MetaMathQA} \citep{yu2023metamath}: A reasoning-focused dataset generated via question bootstrapping, where original math problems are transformed through forward/backward reasoning and rephrasing. It contains 39,500 training samples and is designed to enhance generalization in mathematical reasoning.

    \item \texttt{WizardCoder} \citep{luo2023wizardcoder}: A code generation dataset evolved from Code Alpaca using the Evol-Instruct methodology. It applies iterative complexity-increasing transformations—such as adding constraints, increasing reasoning depth, and introducing erroneous code—to generate approximately 78,000 high-quality programming tasks, enabling robust training for code synthesis models.

    \item \texttt{MBPP} \citep{austin2021program}: A benchmark of around 1,000 Python programming problems written by crowdworkers, targeting entry-level coding skills. Each problem includes a natural language description, reference solution, and three test cases. A subset has been manually verified, ensuring high correctness and usability for evaluation.
\end{itemize}

\subsection{Training Details}
All experiments are conducted on 4 NVIDIA A100 80GB GPUs equipped with an Intel(R) Xeon(R) Platinum 8350C CPU, ensuring consistent hardware conditions across training runs.

\subsubsection{Instruction-following Experiments.}
We use the \texttt{databricks-dolly-15K} dataset \citep{DatabricksBlog2023DollyV2}, which consists of 15,000 human-authored instruction-response pairs. To maintain compatibility with model input constraints, samples exceeding the maximum context length are filtered out. The dataset is randomly partitioned into 12.5K training, 1K validation, and 0.5k test instances.

Hyperparameter selection—including learning rates from \{5e-4, 1e-4, 5e-5\} and batch sizes from \{8, 16\}—is performed via grid search based on validation performance. A key feature of our SRD framework is its training efficiency: the total number of training steps in SRD is reduced to 60\% of that used for baseline models. Specifically, while baseline models are trained for 20 epochs, our method applies 8 epochs per curriculum stage under identical optimization settings, achieving significant training acceleration with reduced computational cost while consistently exceeding the performance of fully trained baselines.

\subsubsection{Task-specific Experiments.}
For text summarization, we use the \texttt{XSum} dataset \citep{gliwa2019samsum}. Both teacher and student models in the baseline setting are trained for 10 epochs. In SRD, the total training steps are again reduced to 60\% of the baseline, maintaining fair comparison while emphasizing data efficiency.

For machine translation, we evaluate on the IWSLT 2017 English-to-German (En-De) dataset \citep{cettolo2017overview}. Given the complexity and smaller scale of this task, we utilize the full dataset without downsampling. The baseline models are trained for 2 epochs with a fixed learning rate of $1 \times 10^{-4}$, and the largest feasible batch size is selected from \{8, 32, 64\} based on GPU memory constraints. All models undergo ten validation checkpoints during training to ensure stable performance monitoring.

In mathematical reasoning and code generation tasks, we build upon models fine-tuned on the \texttt{MetaMathQA} dataset \citep{yu2023metamath} for 100K steps. To elicit step-by-step reasoning, we adopt chain-of-thought (CoT) prompting \citep{wei2022chain} with the instruction: \textit{"Please reason step by step, and put your final answer within \texttt{\textbackslash{}boxed\{\}}."} Baseline models are trained for 2 epochs with a fixed learning rate of $5 \times 10^{-5}$. For all models, we maximize the per-GPU batch size under the constraint of 4 A100 GPUs, adjusting gradient accumulation steps to maintain an effective batch size of 128.

Notably, for instruction-following (OpenLLaMA2), mathematical reasoning, and code generation tasks, we apply LoRA (Low-Rank Adaptation) \citep{hu2022lora}, a parameter-efficient fine-tuning method, to reduce trainable parameters and accelerate training.

\subsubsection{Hyperparameter Settings.}
Across all tasks, we adopt a unified configuration for SRD hyperparameters. The data retention ratio $\lambda$ is set to 0.75, indicating that 75\% of high-quality samples are preserved in each curation stage. The distillation process is divided into $n = 3$ curriculum stages to progressively refine the training data. The temperature parameter $\tau$ is linearly increased from $\tau_0 = 1$ to $\tau_n = 2$, encouraging smoother output distribution alignment in early stages and sharper mimicry in later stages.

For off-policy KD, the ratio $\alpha$ of the SFT loss is linearly decayed from $\alpha_0 = 0.3$ to $\alpha_n = 0.1$, gradually shifting from ground-truth reliance to teacher-guided learning. In contrast, for on-policy KD, no SFT component is used ($\alpha = 0$ throughout), as the student learns exclusively from its own generated outputs refined through teacher feedback. Besides, we follow the optimal mixing strategy proposed by \citet{agarwal2024policy}, using a balanced combination of 50\% SGOs and 50\% ground-truth responses to stabilize training and enhance generalization.

Final model checkpoints are selected based on Rouge-L scores on the validation set, which have been shown to correlate strongly with human evaluation in instruction-following and summarization tasks \citep{agarwal2024policy}.

\subsection{Evaluation Details}
All evaluations are performed on a single NVIDIA A100 80GB GPU, following the protocol in \citet{gu2023minillm}.

\subsubsection{Instruction-following.}
\begin{figure}[!ht]
   \begin{center}
     \includegraphics[scale = 0.1]{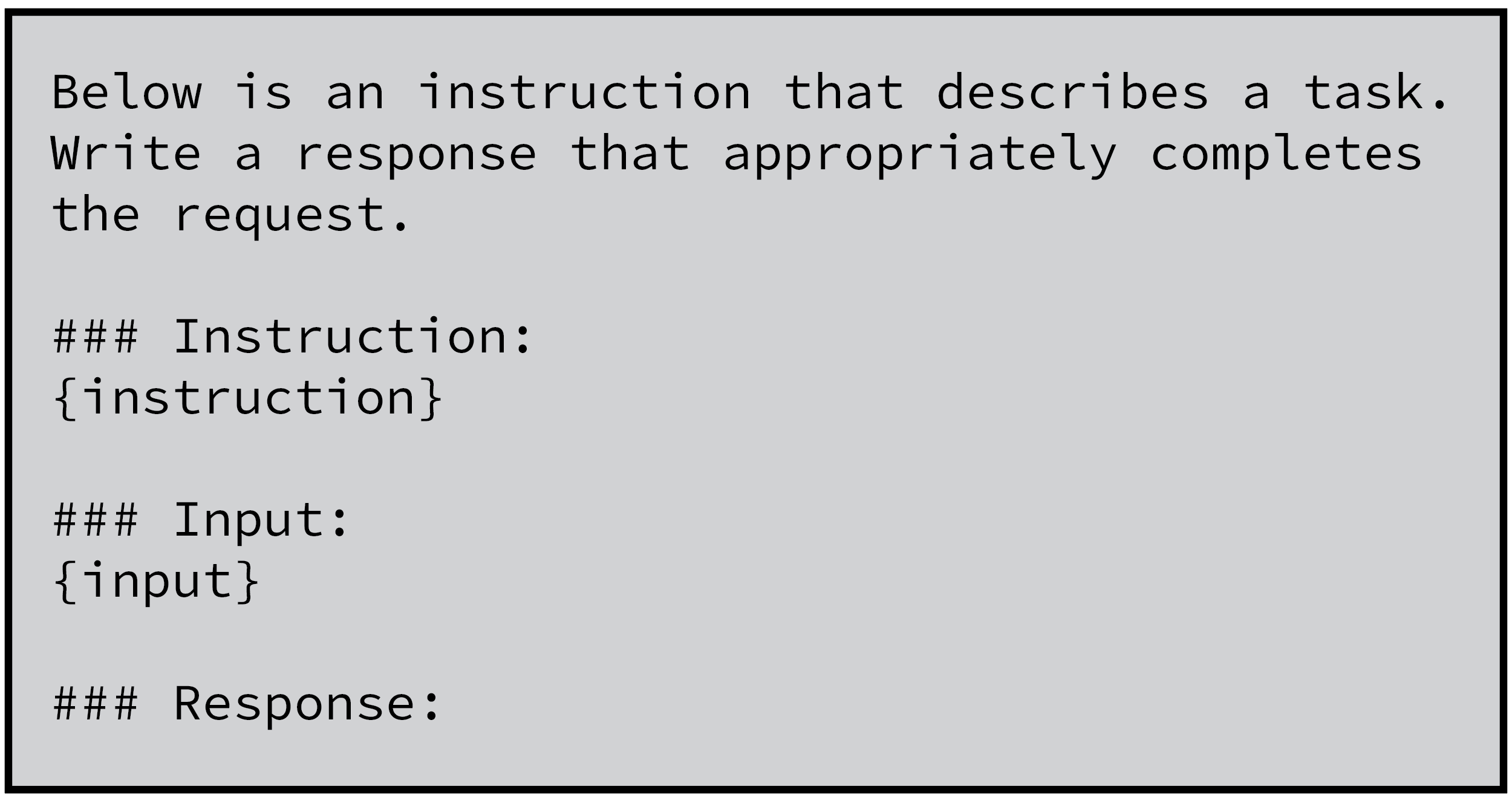}
   \caption{\small Prompt template used in instruction-following experiments, adapted from \citep{gu2023minillm}}\label{fig:prompt}
   \end{center}
\end{figure}

A fixed prompt template (Fig.~\ref{fig:prompt}) is applied during inference to ensure consistency across models and prevent formatting bias. Responses are generated using a temperature of 1.0 and a maximum sequence length of 512 tokens. To improve reliability, we generate five responses per input using different random seeds (\{10, 20, 30, 40, 50\}) and report averaged metrics across these samples.

\subsubsection{Text Summarization and Machine Translation.}
During evaluation, outputs are generated using greedy decoding with a maximum length of 128 tokens. We use ROUGE-L \citep{lin2004rouge} as the primary metric for text summarization and BLEU \citep{papineni2002bleu} for machine translation, both of which are standard and well-established in their respective domains.

\subsubsection{Mathematical Reasoning and Code Generation.}
Responses are sampled using greedy decoding with a maximum generation length of 1024 tokens to accommodate long reasoning chains. For code generation, we employ the EvalPlus framework \citep{liu2023your}, which enhances the original HumanEval benchmark with additional test cases and automated verification, providing a more rigorous and reliable assessment of functional correctness.

\subsection{Baseline Description}
This section outlines the KD loss functions evaluated in our experiments. These losses quantify the discrepancy between the output distribution of the teacher model $p$ and the student model $q_{\theta}$ using various statistical divergence measures. We consider several widely adopted and recently proposed divergence-based objectives to ensure a comprehensive comparison.

The KLD \citep{hinton2015distilling}, a standard choice in KD, is defined as:

\begin{equation}\small\label{eq.07}
    D_{KLD}(p \parallel q_{\theta}) = \mathbb{E}_{y \sim p} \left[\log \frac{p(y)}{q_{\theta}(y)} \right],
\end{equation}

where the expectation is taken over token predictions drawn from the teacher’s distribution $p$. This formulation encourages the student to assign high probability to outputs favored by the teacher.

The Reverse KLD (RKL), which reverses the argument order, is given by:

\begin{equation}\small\label{eq.08}
    D_{RKL}(q_{\theta} \parallel p) = \mathbb{E}_{y \sim p} \left[\log \frac{q_{\theta}(y)}{p(y)} \right].
\end{equation}

Unlike KLD, RKL emphasizes fitting the student distribution to the teacher's high-probability regions and has been shown to improve stability in certain distillation settings.

The Jensen–Shannon Divergence (JSD) \citep{agarwal2024policy} provides a symmetric and smoothed measure of divergence:

\begin{equation}\small\label{eq.09}
    D_{JSD}(p, q_{\theta}) = \frac{1}{2} \mathbb{E}_{y \sim p} \left[\log \frac{p(y)}{m(y)} \right] + \frac{1}{2} \mathbb{E}_{y \sim q_{\theta}} \left[\log \frac{q_{\theta}(y)}{m(y)} \right],
\end{equation}

where $m(y) = \frac{1}{2}p(y) + \frac{1}{2}q_{\theta}(y)$ is the midpoint distribution. JSD balances bidirectional alignment and is less sensitive to distribution mismatches than asymmetric divergences.

The Total Variation Distance (TVD) \citep{wen2023f} measures the $\ell_1$ difference between the two distributions:

\begin{equation}\small\label{eq.10}
    D_{TVD}(p, q_{\theta}) = \frac{1}{2}\sum_{y}|p(y) - q_{\theta}(y)|.
\end{equation}

As a metric bounded in $[0,1]$, TVD offers intuitive interpretability and robustness to outliers, though it lacks gradient signals in regions of non-overlapping support.

The Forward Skew KLD (SKL) \citep{ko2024distillm} introduces a convex combination of teacher and student distributions in the reference argument:

\begin{equation}\small\label{eq.11}
    D_{SKL}(p \parallel q_{\theta}) = D_{KLD}\left(p \parallel \beta p + (1-\beta) q_{\theta} \right),
\end{equation}

while the Reverse Skew KLD (SRKL) is defined as:

\begin{equation}\small\label{eq.12}
    D_{SRKL}(q_{\theta} \parallel p) = D_{KLD}\left(q_{\theta} \parallel (1 - \beta) p + \beta q_{\theta} \right),
\end{equation}

where $\beta \in [0,1]$ controls the interpolation weight. These skewed variants mitigate extreme gradients by smoothing the target distribution, enhancing training stability.

Collectively, these divergence functions represent a diverse set of strategies for aligning student and teacher models in white-box KD, covering symmetric, asymmetric, and regularized forms of distribution matching. Their inclusion demonstrates the flexibility of the proposed SRD framework in accommodating diverse KD loss functions, while consistently achieving strong performance across different divergence measures, underscoring its robustness and adaptability in practical distillation scenarios.

\section{Experimental Results Explanation}
In our experiments, the SRD framework operates on 75\% of the full training data, in contrast to baseline methods that utilize 100\% of the data. This subset is partitioned into three disjoint curriculum stages, each introducing an additional 25\% of the total data. To illustrate, consider the instruction-following task (see Table~1): baseline models are trained for 20 epochs over the complete dataset, equivalent to each training sample being seen 20 times.

In SRD, training proceeds in three stages. In the first stage, only the initial 25\% subset is used and trained for 8 epochs. In the second stage, a second subset is added, resulting in 50\% of the data being trained for another 8 epochs. In the third stage, a third subset is incorporated, bringing the total to 75\% of the data, again trained for 8 epochs. 

This design reduces the total number of training steps in SRD to 60\% of those required by the full-data baseline. Additionally, SRD includes an extra data curation step to evaluate, rank, and select high-quality samples for the next curriculum phase. This reflection step typically accounts for 5\%–7\% of the total runtime of the corresponding baseline, making it computationally negligible relative to overall savings.

To quantify the time efficiency, we measure wall-clock training time on identical hardware: four NVIDIA A100 80GB GPUs and an Intel(R) Xeon(R) Platinum 8350C CPU. For the instruction-following task using the OpenLLaMA2-3B student model, the data curation step—evaluating and ranking all 11,435 training samples from \texttt{databricks-dolly-15k}—takes approximately 14 minutes.

Comparing total training time:

\begin{itemize}
    \item \textbf{Offline KD baseline}: $\sim$ 290 minutes of training $\rightarrow$ SRD total time: $14 + 0.6 \times 290 \approx 188$ minutes, yielding a \textbf{35.18\%} reduction in training time.
    \item \textbf{Online/on-policy KD baseline}: $\sim$ 1500 minutes of training $\rightarrow$ SRD total time: $14 + 0.6 \times 1500 \approx 914$ minutes, resulting in a \textbf{39.07\%} time saving.
\end{itemize}

These results demonstrate that SRD achieves substantial computational efficiency without sacrificing model performance, as validated in the main experiments. 

\section{Additional Related Works}
White-box KD for LLMs leverages soft labels from teacher models to outperform black-box KD \citep{yang2024survey}, with prior work focusing on divergence functions such as KLD \citep{hinton2015distilling}, RKL \citep{gu2023minillm}, JSD \citep{agarwal2024policy}, SKL and SRKL \citep{ko2024distillm}, and $\alpha$-$\beta$-Divergence \citep{wang2025abkd}. Complementary data curation strategies incorporate TGOs/SGOs \citep{kim2016sequence, agarwal2024policy} to improve training dynamics, yet often overlook the critical role of data quality and student-model compatibility—a principle well-established in black-box KD and SFT \citep{li2023quantity, li2024selective, ding2023enhancing}. For instance, UltraChat \citep{ding2023enhancing} curates diverse, high-quality data for black-box KD, while phi \citep{gunasekar2023textbooks} emphasizes distillation from compact, expert-curated datasets.  

Curriculum learning (CL) has been explored to address when data should be introduced during training, though primarily in vision \citep{xiang2020learning, li2023curriculum} and limited NLP applications \citep{zhu2021combining}. Recent efforts include MPDistil \citep{sengupta2023good}, which uses reinforcement-based curriculum scheduling, Confucius \citep{gao2024confucius} for tool learning via black-box KD, and POCL \citep{liu2025being}, a plug-in framework inspired by "progressive overload." However, these methods largely neglect the interplay between sample difficulty, data quality, and student capacity—a key gap in current white-box KD.

\section{Limitations}
In mathematical reasoning and code generation tasks, SRD achieves positive but relatively modest performance gains compared to its effectiveness in other domains. This suggests a reduced impact of the current data curation strategy in settings where output correctness is highly sensitive to small errors—such as arithmetic miscalculations or syntactic inaccuracies—that are not reliably detected by standard similarity metrics like ROUGE-L or token-level cross-entropy used in the Selective Reflection on Training Data stage. These metrics may assign high scores to outputs that are superficially similar to reference solutions but functionally incorrect.

This limitation highlights the need for task-adaptive difficulty assessment mechanisms in KD, particularly for domains requiring precise semantic or executable correctness. Future work should explore integrating execution-based evaluation—such as code compilation, test case execution, or mathematical answer verification—into the data curation pipeline. Additionally, incorporating semantic equivalence scoring, which accounts for correct solutions expressed in different forms, could improve the fidelity of sample selection. Such enhancements would enable SRD to better identify and prioritize high-quality, functionally correct responses, thereby improving distillation efficacy in rigorous reasoning and generation tasks.

\end{document}